\documentclass[sigconf, nonacm]{acmart}

\usepackage{amsmath,amssymb}
\usepackage{amsmath,amssymb}
\usepackage{graphicx}
\usepackage{multirow}
\usepackage{microtype}
\usepackage{xcolor}
\usepackage{lipsum}
\usepackage[linesnumbered,ruled,vlined,noend]{algorithm2e}
\usepackage{pifont}
\usepackage{soul}
\usepackage{tabularx}
\usepackage{tikz}
\usetikzlibrary{shapes.geometric,shapes.misc,positioning,calc,fit,backgrounds,arrows.meta}
\newcommand{\cmark}{\ding{51}}
\newcommand{\xmark}{\ding{55}}
\newcommand{\myNum}[1]{(\emph{#1})}

\newcommand{\ie}{{\it i.e.}, }

\def\GraphCast{GraphCast}
\def\Aurora{Aurora}
\def\CRA5{CRA5}
\def\SZ{SZ3.1}

\definecolor{hldonegreen}{rgb}{0.62,0.94,0.62}

\SetKwComment{Comment}{{// }}{}
\AtBeginDocument{%
  }

\renewcommand\footnotetextcopyrightpermission[1]{}
\begin{document}

\title{Can Deep Neural Networks Improve Compression of Very Large Scientific Data?}

\author{Muhannad Alhumaidi}
\affiliation{%
  \institution{King Abdullah University of Science and Technology}
  \city{Thuwal}
  \country{Saudi Arabia}}
\email{muhannad.humaidi@kaust.edu.sa}

\author{Guozhong Li}
\affiliation{%
  \institution{King Abdullah University of Science and Technology}
  \city{Thuwal}
  \country{Saudi Arabia}}
\email{guozhong.li@kaust.edu.sa}

\author{Spiros Skiadopoulos}
\affiliation{%
  \institution{University of the Peloponnese}
  \city{Tripoli}
  \country{Greece}}
\email{spiros@uop.gr}

\author{Panos Kalnis}
\affiliation{%
  \institution{King Abdullah University of Science and Technology}
  \city{Thuwal}
  \country{Saudi Arabia}}
\email{panos.kalnis@kaust.edu.sa}

\renewcommand{\shortauthors}{Alhumaidi et al.}

\begin{abstract}

Error-bounded lossy compression is a fundamental technique for managing the rapidly growing volumes of scientific data produced by modern simulations and observational instruments. Most state-of-the-art compressors follow a prediction–residual paradigm, where compression effectiveness depends on the quality of the predictor: more accurate predictions generate smaller residuals that are easier to compress. This observation raises a question: can modern machine learning models serve as superior predictors for scientific data compression?
Answering this question directly is challenging because developing compression-specific ML predictors requires substantial resources. Instead, we leverage the climate domain where highly accurate pretrained weather forecasting foundation models
already exist, making them an ideal testbed.
We present a framework that integrates spatial and temporal deep learning models into a conventional error-bounded compression pipeline. The framework supports auto-regressive forecasting models and avoids error accumulation. Using ERA5 climate data as a representative large-scale scientific dataset, we evaluate three distinct ML predictors: a VAEformer-based codec (CRA5), a graph neural network forecaster (GraphCast), and a vision-transformer forecaster (Aurora), against the state-of-the-art compressor SZ3.1 under identical quantization and entropy-coding backends.
Our evaluation over approximately 1.7 TB of data reveals a surprising result: although ML predictors generate  more accurate predictions and can improve reconstruction quality by up to 91\% while achieving up to 9.6$\times$ higher compression ratios for highly predictable variables, they do not improve overall dataset-level compression ratio. We show that prediction accuracy alone is insufficient: the spatial structure of the resulting residuals plays a decisive role in entropy coding efficiency. Our findings provide new insight into the relationship between prediction and compression, establishing the opportunities and limitations of using foundation models as predictors for scientific data compression.
\end{abstract}

\maketitle

\section{Introduction}
\label{sec:intro}
Modern sensing and simulation generate scientific datasets at the terabyte to petabyte scale across many domains, including weather and climate modeling~\cite{hoteit-RSRA2018,kay2015community,hoteit-RSRA2022}, large-scale reanalysis~\cite{hersbach2020era5}, cosmological simulations~\cite{nyx_simulation}, computational fluid dynamics and turbulence~\cite{miranda_application,li2008public}, combustion~\cite{chung2022blastnet}, and seismic imaging~\cite{kayum2020geodrive}. At these volumes, storage, network, and compute infrastructures become the dominant bottleneck~\cite{gray2005scientific}, making data \emph{compression} essential for tractable storage, transfer, and analysis.

\emph{Lossless} compression, as implemented in scientific data libraries (e.g., NetCDF~\cite{rew1990netcdf}, HDF5\footnote{\url{http://www.hdfgroup.org/HDF5}}) through algorithms like  Zlib\footnote{\url{http://www.zlib.net/}} and  Zstandard\footnote{\url{https://github.com/facebook/zstd/}},  
recovers data exactly but compression ratios rarely 
exceed a factor of $2$~\cite{zhao2020sdrbench}. In contrast, \emph{lossy} 
compression~\cite{klema1980singular,jolliffe2016principal,huang2023compressing} can push 
compression ratios to thousands, but without any control over the errors the decompressed 
data may be unsuitable for scientific use. This  led the scientific community toward a middle ground: \emph{error-bounded lossy} compression. The idea is to allow some loss of precision, but to guarantee that a 
user-chosen error metric, such as relative point-wise error, to never exceed a 
predefined bound $\epsilon$.

\begin{figure}
    \centering
    \includegraphics[width=0.8\linewidth]{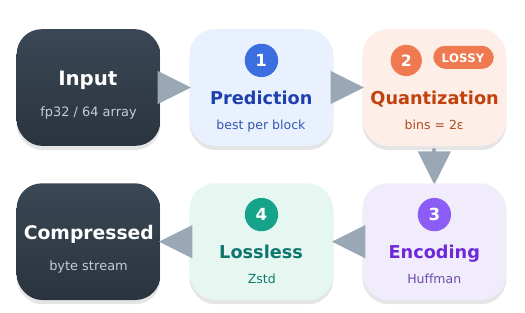}
    \caption{Typical (e.g., SZ3.1~\cite{zhao2021optimizing}) error-bounded lossy compression pipeline. We focus on the \textbf{Prediction} module.}
    \label{fig:sz3_pipeline}
\end{figure}

Many error-bounded lossy compression methods are built around the \emph{prediction-residual} architecture depicted in Figure~\ref{fig:sz3_pipeline}. A prediction module approximates input data\footnote{Scientific data encompasses various types, including numerical, categorical, text, and image data; this paper focuses on \emph{numerical} data.} points;  the error-bounding module then quantizes and encodes only the residual between that prediction and the true value.
Existing compressors approach the prediction problem in several  ways: HPEZ~\cite{liu2023high} and SZ3.1~\cite{zhao2021optimizing} combine multiple interpolation schemes to predict from spatially neighboring points, 
while SPERR~\cite{li2023lossy} and ZFP~\cite{lindstrom2014fixed} operate in a wavelet-transformed domain. Recently, deep neural networks have entered the picture, giving rise to autoencoder-based SZ~\cite{liu2021exploring}, coordinate 
network-based compressors~\cite{han2022coordnet}, and super-resolution network-based SZ~\cite{liu2023srn}. 

The  pipeline of Figure~\ref{fig:sz3_pipeline} carries a  
consequential implication: the quality of the predictor determines
the  compression ratio. A good predictor keeps residuals small and tightly clustered, 
producing a data stream that entropy coders can compress aggressively. 

This observation raises the central question of our work: \emph{if the simple, general-purpose predictor at the heart of these compressors were replaced by a highly elaborate, ML-trained model, would it serve as a more effective prediction front-end and yield a better compression pipeline?}
Given the centrality of predictor quality to compression ratio, a learned model that closely tracks the data, i.e., one able to exploit physical priors and long-range spatiotemporal dependencies that are invisible to polynomial or wavelet-based predictors, should in principle produce residuals of substantially lower variance, directly translating to higher compressibility.

However, training and validating such a model from scratch is a significant task. It would require large amounts of training data, expensive computational budgets, careful architectural design, and extensive empirical validation. Such resources  would need to be justified before the viability of the compression use case is even established.

A pragmatic and scientifically sound alternative is to focus on a domain where highly accurate ML models \emph{already exist} and to repurpose these \emph{pretrained} models. Climate is exactly such a domain. 
Recent advances in data-driven weather prediction have produced models such as 
GraphCast~\cite{lam2023learning}, 
Aurora~\cite{bodnar2024aurora}, 
Pangu-Weather~\cite{bi2023pangu}, 
and FourCastNet~\cite{pathak2022fourcastnet}. 
The models are trained on the massive ERA5~\cite{hersbach2020era5} global reanalysis dataset produced by the  European Centre for Medium-Range Weather Forecasts (ECMWF), which provides hourly estimates of a wide range of atmospheric, land-surface, and oceanic variables at a spatial resolution of $0.25^{\circ}$ from 1940 to the present. 
The ML models already implicitly encode the spatiotemporal structure of ERA5 in their learned representations. 
Using them as predictors within a prediction-residual compression framework is therefore not merely a shortcut 
but a principled proxy experiment: their performance on the compression task will give us a reliable signal about 
whether deep learning-based prediction is a viable direction for scientific data compression, 
and will expose the bottlenecks that any such approach, including purpose-trained models, would need to overcome.

In this paper, we present a systematic evaluation of the prediction-residual
compression framework applied to ERA5 data, comparing three qualitatively
different deep learning-based prediction strategies against an established baseline, within a shared error-bounded back-end.
\myNum{i} \CRA5~\cite{han2023cra5} is a \emph{learned spatial} predictor based on a VAEformer encoder-decoder that reconstructs each variable independently, without conditioning on adjacent timesteps. It was specifically trained to compress ERA5 data without error bound. It can be integrated in the error-bounded pipeline of Figure~\ref{fig:sz3_pipeline} in a straight-forward manner, by replacing the Prediction module.    
\myNum{ii} GraphCast~\cite{lam2023learning} is a mesh-based graph neural network, that serves as \emph{temporal} predictor by receiving the current state as input and predicting the next few states. To integrate GraphCast in the error-bounded compression framework, we need to modify the pipeline in an \emph{auto-regressive} way: we route each prediction through the error-bounding module, where the residual is calculated. The residual is combined with the prediction to generate a ``corrected'' current state that is used as the next input for the predictor; refer to Section~\ref{ssec:models} for details.   
\myNum{iii} Aurora~\cite{bodnar2024aurora} is a patch-based vision transformer that also serves as temporal predictor. Similar to GraphCast, in order to ensure error-bounded compression, we integrate Aurora within our auto-regressive compression pipeline. 
\myNum{iv} SZ3.1~\cite{zhao2021optimizing} serves as the classical \emph{spatial polynomial} baseline. All four methods share an identical quantization and entropy coding back-end, for fair comparison of each prediction strategy.

Our evaluation spans 1{,}997 autoregressive steps ($\sim$500 days) across 9~atmospheric variables at three relative error bounds $\epsilon \in \{10^{-2}, 10^{-3}, 10^{-4}\}$. Our results reveal that, when considering the dataset as a whole, none of the ML-based methods can match the compression ratio of SZ3.1, because their residuals, while smaller in magnitude, lack the spatial coherence that entropy coding exploits. Interestingly, if we consider each variable separately, for some highly-predicatble variables the ML-based methods achieve up to $9.6\times$ higher compression ratios than SZ3.1. Note that, even though the point-wise error is bounded for all methods, the fidelity of the decompressed results is different: SZ3.1 generally results in larger average error, and the error distribution tends to generate structured artifacts that are undesirable for scientific analysis. In contrast the mean absolute error (MAE) of CRA5 and Aurora is up to 91\% and 34\% lower than SZ3.1, respectively, while the error distribution is much smoother. Interestingly, the auto-regressive design for Aurora and GraphCast is stable over nearly 2,000 steps with no error accumulation.      
In summary, the answer to our main question is nuanced: when used as predictors in an error-bounded lossy compression pipeline, these elaborate ML models can dramatically improve reconstruction fidelity, yet do not by themselves yield higher compression ratios.

Our contributions are as follows:
\begin{itemize}
    \item We develop an auto-regressive framework that enables temporal forecasting models to be used as predictors in an error-bounded lossy compression pipeline, without error accumulation.
    
    \item We integrate three representative elaborate ML models, a VAEformer (CRA5), a mesh-based GNN (GraphCast) and a patch-based vision transformer (Aurora), within an identical error-bounded quantization and entropy coding
    back-end, enabling a fair comparison of prediction strategies for ERA5 compression.

    \item We present a comprehensive large-scale (around 1.7TB of data) experimental evaluation with analysis per-variable,
    per-level, and per-pixel. We show that ML-based predictors can significantly improve reconstruction fidelity and may even improve compression ratio for some variables; however, they cannot match the compression ratio of the simpler predictors of SZ3.1 for the entire ERA5 dataset.   
\end{itemize}

\section{Background}
\label{sec:background}

\subsection{Error-bounded lossy compression}

A temporal scientific dataset $\mathcal{X}$ captures data across multiple timestamps, organized in a set of \emph{variables} such as temperature, humidity, etc. Each variable $X \in \mathcal{X}$ is a multi-dimensional matrix of real numbers. For the rest of this paper we assume that $X \in \mathbb{R}^{T \times M \times N}$, where $T$ is the number of timesteps and $M,N$ are the number of rows and columns, respectively. 

For a variable $X$, let $|X|$ be the size of the original data in bytes. Consider some compression method, and let $|Z|$ be the total size
of the compressed data in bytes; this includes all the \emph{overhead} data needed for
decompression, such as metadata, initialization values for auto-regressive predictors, learned latent representations, ML model weights, etc. The compression ratio $\rho$ is defined as:
\begin{equation}
\rho = \frac{|X|}{|Z|}
\end{equation}
For the entire dataset $\mathcal{X}$, the formula is generalized to $\rho = \frac{\mathcal{|X|}}{\mathcal{|Z|}}$. 
In case part of the overhead (e.g., the ML model weights) is considered  common knowledge, for fairness we will exclude that part from the calculation of $|Z|$ and denote compression ratio as $\overline{\rho}$.

\paragraph{Error-bounded lossy compression.}
For a variable $X$, let $\epsilon$ be a user-specified \emph{relative} error bound. Let $C$ be a compression
function that generates the compressed version $Z$ of $X$, \ie $Z = C(X)$, and
let $D$ be a decompression function that generates the decompressed version $X'$
of $Z$, \ie $X' = D(Z)$. The process is lossy, therefore $X'$ is an approximation
of $X$. The goal is to define a pair of functions $\langle C, D \rangle$ such that
the compression ratio $\rho$ of $X$ is maximized, while the relative difference
between each original data point $X_i$ and its decompressed counterpart
$X'_i = D(C(X))_i$ does not exceed the error bound $\epsilon$. Formally:
\begin{equation}
\operatorname*{arg\,max}_{\langle C, D \rangle} \frac{|X|}{|C(X)|},
\quad \text{such that} \quad
\frac{|X_i - D(C(X))_i|}{r_X} \leq \epsilon
\end{equation}
where $r_X$ returns the range of values in the variable.

\subsection{Prediction-residual compression pipeline}

Many error-bounded lossy compression methods \cite{han2022coordnet, li2023lossy, lindstrom2014fixed, liu2023srn, liu2023high, liu2021exploring, zhao2021optimizing} follow a pipeline similar to that of Figure~\ref{fig:sz3_pipeline}, their main differences being the way they prepare the input data (i.e., step-0) and their prediction method (i.e., step-1). For the rest of this paper we will focus on \SZ{}~\cite{zhao2021optimizing}, which is a popular state-of-the-art general-purpose method. We will use \SZ{} as a representative high-performance baseline, with the caveat that, for specific application domains, some other method with a particular optimization heuristic may perform slightly better.

\begin{figure}[t]
    \centering
    \includegraphics[width=1\linewidth]{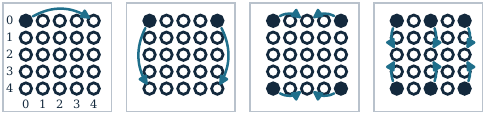}
    \caption{Polynomial prediction example. Black and white circles represent processed and unprocessed data points, respectively; arrows denote interpolation.}
    \label{fig:prediction_example}
\end{figure}

\SZ{}  uses a  \emph{spatial} predictor for step-1 of Figure~\ref{fig:sz3_pipeline}: it fits a polynomial interpolant (see Figure~\ref{fig:prediction_example}) to neighboring
points within the same timestep $k$ to predict values $p_{X,k}$ of variable $X$, computes the residual $d_{X,k}$
against the true values $t_{X,k}$, quantizes it under error bound $\epsilon$, and
entropy-codes the result. \SZ{} has no access to temporal information; each
timestep is compressed independently. The residual is computed as:
\begin{equation}
    d_{X,k} = t_{X,k} - p_{X,k}
\end{equation}

\paragraph{Error-bounded quantization (step-2).}
To enforce a relative point-wise error bound $\epsilon$, the quantization bin width
is defined as a function of $\epsilon$ and the global value range of each
variable. Due to quantization, this is the \emph{only lossy step in the pipeline}.
For a variable $X$, the global minimum and maximum are computed in advance across all timesteps $1 \leq k \leq T$ and spatial locations $1 \leq i \leq M$, $1 \leq j \leq N$; then the range is:
\begin{equation}
    r_{X} = \max_{k,i,j}\, t_{X,k}^{(i,j)} - \min_{k,i,j}\, t_{X,k}^{(i,j)}
\end{equation}
The quantization bin width is set to:
\begin{equation}
    \delta_{X} = \epsilon \cdot r_{X}
\end{equation}
This bin width is fixed for the entire compression sequence; it does not vary across
timesteps. Each residual value is then mapped to an integer bin index:
\begin{equation}
    q_{X,k}^{(i,j)} = \left\lfloor \frac{d_{X,k}^{(i,j)}}{\delta_{X}} + \frac{1}{2} \right\rfloor , \quad \text{where} \quad q_{X,k} \in \mathbb{Z}^{M \times N} 
\end{equation}

\paragraph{Entropy coding (step-3) and compression (step-4).}
Huffman encoding \cite{Huffman1952} compresses data \emph{losslessly} by assigning shorter binary codes to frequently occurring characters and longer codes to rare ones. In step-3 the \SZ{} pipeline uses the Zlib\footnote{\url{http://www.zlib.net/}} implementation of Huffman encoding to compress the integer quantized residuals $q_{X,k}$. Assuming that $p_{X,k}$ is an accurate prediction of $t_{X,k}$, the
residuals $d_{X,k}$ are small in magnitude, and the corresponding bin indices $q_{X,k}$ cluster around zero. This produces a highly skewed integer distribution that Huffman coding can exploit aggressively.

The above mentioned quantization pipeline excludes outliers which are marked as ``unpredictable''. The true values of those outliers
are combined with the output of the Huffman encoder and are further compressed \emph{losslessly} in step-4 by the Zstd\footnote{\url{https://github.com/facebook/zstd/}} algorithm.

\subsection{Decompression}
Decompression follows the inverse pipeline of Figure~\ref{fig:sz3_pipeline}. The crucial step is dequantization, which recovers an approximation of the residual:
\begin{equation}
    \tilde{d}_{X,k}^{(i,j)} = q_{X,k}^{(i,j)} \cdot \delta_{X}
\end{equation}
The error-corrected state is then:
\begin{equation}
    p'_{X,k} = p_{X,k} + \tilde{d}_{X,k}
\end{equation}
which satisfies the relative error bound by construction:
\begin{equation}
    \left| p_{X,k}^{\prime(i,j)} - t_{X,k}^{(i,j)} \right| \leq \frac{\delta_{X}}{2} = \frac{\epsilon \cdot r_{X}}{2}.
\end{equation}

\section{Methodology}\label{sec:method}

Section~\ref{sec:background} established that, in the prediction--residual
pipeline, the predictor is the component that determines the compression
ratio: predictions that are accurate produce residuals that are small in
magnitude and tightly clustered around zero, leaving the entropy coder little
to encode. The quality of the predictor is therefore crucial for
error-bounded lossy compression of scientific data. Classical compressors such
as \SZ{} rely on simple, general-purpose predictors (e.g., spatial polynomial
interpolation) that are cheap to evaluate but blind to the physical priors and
long-range spatiotemporal dependencies that govern scientific fields.
Our work aims to investigate whether an elaborate, ML-trained model can act as a
more accurate predictor to improve the compression pipeline.
 
Answering this question for scientific data \emph{in general} is not practical.
A predictor purpose-built for compression would require an excessive amount of resources that are beyond the reach of most academic environments. 
Rather than train such a model from scratch, we adopt a pragmatic and
scientifically sound alternative: we restrict attention to a single domain in
which highly accurate ML models \emph{already exist} and re-purpose them as
predictors behind a fixed error-bounded back-end. Below we argue that climate is an appropriate target for such an investigation for two reasons: there exists a plethora of highly-curated open source data, and there exist several accurate pre-trained ML models adopted by the scientific community. 

\paragraph{ERA5 dataset.} 
\label{ssec:era5}

\begin{figure}[t]
    \centering
    \includegraphics[width=\columnwidth]{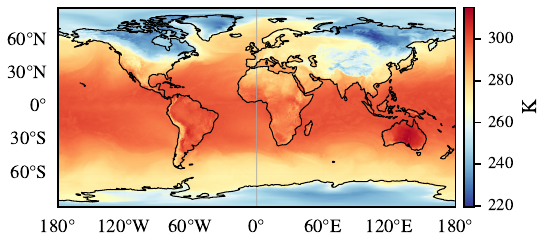}
    \caption{A single ERA5 field: 2-meter temperature (in $K$) at one timestep, on the
    $0.25^{\circ}$ ($721 \times 1440$) latitude-longitude grid. The field has clear
    large-scale spatial structure, warm in the tropics and colder toward the poles. 
    }
    \label{fig:field_map}
\end{figure}

ERA5~\cite{hersbach2020era5} is a global climate reanalysis dataset that combines past weather observations with advanced numerical models to create a continuous record of Earth's atmosphere, land, and oceans. It was developed by the European Centre for Medium-Range Weather Forecasts (ECMWF). ERA5 unifies data from satellites, weather balloons, and ground sensors to provide a highly detailed, historically consistent map of global climate conditions dating back to 1940. 
ERA5 contains hourly values of a variety of variables such as temperature, humidity and wind at several pressure levels (i.e., altitudes) in the atmosphere, on a $0.25^{\circ}$
latitude-longitude global grid that corresponds to a $721 \times 1440$ matrix; an example is shown in Figure~\ref{fig:field_map}. A variable at a specific pressure level is a 3D matrix $X \in \mathbb{R}^{T \times M \times N}$ (refer to Section~\ref{sec:background}). The total size of ERA5 exceeds 5 PB, depending on the format; in this study we use an extract of approximately 1.7 TB.

\paragraph{ML-based predictors.}
There exist several pre-trained ML models \cite{bi2023pangu, bodnar2024aurora, han2023cra5, lam2023learning, pathak2022fourcastnet} that forecast accurately the atmospheric conditions.
In this study, we focus on three representative such models: 
{\CRA5}~\cite{han2023cra5} is a learned spatial codec (i.e., VAEformer) built specifically to compress
ERA5. It is the most direct learned competitor to a classical compressor, so it is
the natural learned-spatial point of comparison.
{\GraphCast}~\cite{lam2023learning}, a graph neural network, and {\Aurora}~\cite{bodnar2024aurora}, a vision transformer, are temporal forecasters. We select them for three reasons:
\myNum{i} both are among the best forecasting models at the time of this work; \myNum{ii} both release documented inference code and pretrained weights, so the evaluation is reproducible; and \myNum{iii} 
they are trained on ERA5, so they already encode its spatial structure and temporal dynamics.
We implement all methods using the same quantization and entropy-coding back-end, for fair comparison.

\subsection{\CRA5: Learned Spatial Reconstruction}
\label{ssec:method_cra5}

{\CRA5}~\cite{han2023cra5} is a neural codec built specifically to compress ERA5.
Its design follows learned image compression: a transformer-based variational autoencoder (VAE) called
the VAEformer, built as a dual-VAE. A main VAE maps each atmospheric state to a
quantized latent representation, and a second, hyperprior VAE models the latent distribution and
produces the parameters of a Gaussian entropy model for arithmetic coding. The result
is a reconstruction of each field on its own. {\CRA5} compresses an extract of the  ERA5 archive
(about 226\,TB) to roughly 0.7\,TB, a ratio around $322\times$, but it is
rate-distortion optimized, without any point-wise error guarantee.

\paragraph{Fitting CRA5 into the pipeline.}
CRA5 replaces the Prediction module (i.e., step-1) in the pipeline of Figure~\ref{fig:sz3_pipeline}.
We use {\CRA5} in inference mode with the released weights and no
change to the architecture.
We employ the CRA5 codec to decode each variable $X$ and use that reconstruction as the prediction $p_{X,k}$ for step $k$. The residual $d_{X,k} = t_{X,k} - p_{X,k}$ then passes through the same
error-bounded back-end of Section~\ref{sec:background}: the residual is quantized such that the reconstruction is bounded by relative error $\epsilon$, and then the quantized data are compressed by Huffman followed by Zstd. This process transforms {\CRA5} from an
unbounded codec into an error-bounded one.

Decompression is straight-forward, too: the decoder forms the
{\CRA5} reconstruction and applies the error correction of Section~\ref{sec:background}
to recover $p'_{X,k}$.

\subsection{\GraphCast: Auto-regressive Prediction}
\label{ssec:method_gc}
\label{ssec:models}
{\GraphCast}~\cite{lam2023learning} is a graph neural network with an
encode-process-decode design on a multi-mesh icosahedral grid. It maps the
latitude-longitude ERA5 grid onto the mesh, runs 16 learned message-passing layers that
move information across spatial scales (the finest mesh has 40{,}962 nodes), and maps
the result back to the grid. The model has 36.7\,M parameters and was trained on ERA5
over the period 1979 to 2017. It predicts 5 surface and 6 atmospheric
variables across 37 pressure levels. It takes two consecutive states at $t_{k-2}$ and
$t_{k-1}$ plus static forcing (solar radiation, land-sea mask) and outputs the state at
$t_k$ on the $721 \times 1440$ grid at a 6-hour interval. It forecasts more accurately
than ECMWF's operational high resolution forecast system on about 90\% of the reported targets.

\paragraph{Auto-regressive integration.}
{\GraphCast} cannot be integrated directly into the pipeline of  Figure~\ref{fig:sz3_pipeline}. 
It is a temporal forecaster, so it predicts the next state from the two preceding ones, but at decode
time those preceding states are not available as ground truth. We therefore propose to wrap it in
an auto-regressive closed loop and feed back the decompressed (i.e., ``corrected'') state. 

\begin{figure}[tp]
\centering
\resizebox{\columnwidth}{!}{\begin{tikzpicture}[
  font=\small,
  >=Latex,
  inp/.style ={draw=black!45, line width=0.5pt, rounded corners=3pt, fill=white,
               align=center, minimum width=12mm, minimum height=8mm, inner sep=2.5pt},
  model/.style={draw=orange!75!black, line width=0.5pt, trapezium,
                trapezium left angle=72, trapezium right angle=108,
                fill=orange!22, minimum width=10mm, minimum height=8mm, inner sep=2pt},
  pred/.style ={draw=red!60!black, line width=0.5pt, rounded corners=3pt,
                fill=red!12, minimum height=8mm, minimum width=12mm},
  eb/.style   ={draw=black!50, line width=0.5pt, rounded corners=3pt, fill=black!5,
                align=center, minimum height=8mm, minimum width=20mm, inner sep=2pt},
  corr/.style ={draw=green!50!black, line width=0.5pt, rounded corners=3pt,
                fill=green!18, minimum height=8mm, minimum width=12mm},
  tag/.style  ={draw=black!30, line width=0.4pt, rounded corners=2pt, fill=white,
                font=\footnotesize, inner sep=2pt},
  disk/.style ={draw=black!35, line width=0.4pt, cylinder, shape border rotate=90,
                aspect=0.22, fill=white, font=\footnotesize,
                minimum width=8mm, minimum height=5mm, inner sep=1pt},
  klab/.style ={font=\small, text=black!55},
  flow/.style ={->, line width=0.7pt, black!75},
  side/.style ={->, line width=0.55pt, black!45},
  fb/.style   ={->, line width=0.7pt, blue!60!black},
]
\def\xk{0.1} \def\xa{1.5} \def\xb{3.0} \def\xc{4.5} \def\xd{6.5} \def\xe{8.5}
\def\rA{0} \def\rB{-1.7} \def\rN{-3.4}

\node[klab] at (\xk,\rA) {$k{=}2$};
\node[inp]  (i1) at (\xa,\rA) {$t_{X,0}$\\$t_{X,1}$};
\node[model](m1) at (\xb,\rA) {$\mathcal{F}$};
\node[pred] (p1) at (\xc,\rA) {$p_{X,2}$};
\node[eb]   (e1) at (\xd,\rA) {Error-bound\\[-1pt]{\footnotesize quantize + correct}};
\node[corr] (c1) at (\xe,\rA) {$p'_{X,2}$};
\draw[flow] (i1)--(m1); \draw[flow] (m1)--(p1); \draw[flow] (p1)--(e1); \draw[flow] (e1)--(c1);

\node[klab] at (\xk,\rB) {$k{=}3$};
\node[inp]  (i2) at (\xa,\rB) {$t_{X,1}$\\$p'_{X,2}$};
\node[model](m2) at (\xb,\rB) {$\mathcal{F}$};
\node[pred] (p2) at (\xc,\rB) {$p_{X,3}$};
\node[eb]   (e2) at (\xd,\rB) {Error-bound\\[-1pt]{\footnotesize quantize + correct}};
\node[corr] (c2) at (\xe,\rB) {$p'_{X,3}$};
\draw[flow] (i2)--(m2); \draw[flow] (m2)--(p2); \draw[flow] (p2)--(e2); \draw[flow] (e2)--(c2);

\node[font=\large\bfseries, text=black!70] at (\xc,{(\rB+\rN)/2}) {$\vdots$};

\node[klab] at (\xk,\rN) {$k{=}N{+}1$};
\node[inp]  (iN) at (\xa,\rN) {$p'_{X,N-1}$\\$p'_{X,N}$};
\node[model](mN) at (\xb,\rN) {$\mathcal{F}$};
\node[pred] (pN) at (\xc,\rN) {$p_{X,N+1}$};
\node[eb]   (eN) at (\xd,\rN) {Error-bound\\[-1pt]{\footnotesize quantize + correct}};
\node[corr] (cN) at (\xe,\rN) {$p'_{X,N+1}$};
\draw[flow] (iN)--(mN); \draw[flow] (mN)--(pN); \draw[flow] (pN)--(eN); \draw[flow] (eN)--(cN);

\draw[fb] (c1.east) -- ++(0.35,0) |- (\xa,{(\rA+\rB)/2}) -- (i2.north);
\draw[fb] (c2.east) -- ++(0.35,0) |- (\xa,{(\rB+\rN)/2}) -- (iN.north);

\node[tag]  (g1) at (\xd+0.78,\rA+0.78) {$t_{X,2}$};
\node[disk] (q1) at (\xd-0.78,\rA-0.78) {$q_{X,2}$};
\draw[side] (g1)--(e1); \draw[side] (e1)--(q1);
\node[tag]  (g2) at (\xd+0.78,\rB+0.78) {$t_{X,3}$};
\node[disk] (q2) at (\xd-0.78,\rB-0.78) {$q_{X,3}$};
\draw[side] (g2)--(e2); \draw[side] (e2)--(q2);
\node[tag]  (gN) at (\xd+0.78,\rN+0.78) {$t_{X,N+1}$};
\node[disk] (qN) at (\xd-0.78,\rN-0.78) {$q_{X,N+1}$};
\draw[side] (gN)--(eN); \draw[side] (eN)--(qN);

\end{tikzpicture}}
\caption{Autoregressive loop for the temporal forecasters ({\GraphCast},
 Aurora). Two seed states are stored losslessly. At each step $k$, $\mathcal{F}$
predicts $p_{X,k}$ from the two preceding corrected states; the error-bounding
module stores the quantized residual $q_{X,k}$ and feeds back the corrected state
$p'_{X,k}$, keeping every input within $\epsilon\,r_X/2$ of the truth and
preventing drift.}
\label{fig:gc_pred}
\end{figure}
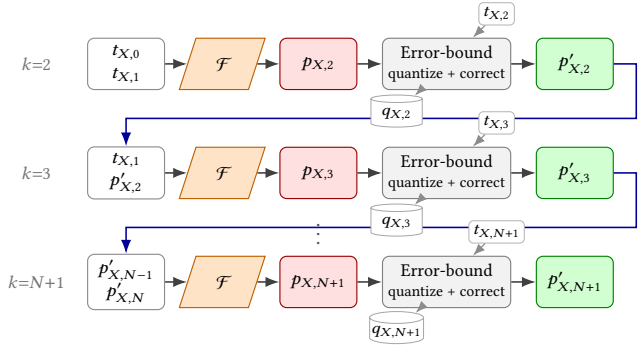

Let $\mathbf{t}_k$, $\mathbf{p}_k$, and $\mathbf{p}'_k$ be the
full multi-variable states at step $k$, that is, the stacks of the per-variable slices
$t_{X,k}$, $p_{X,k}$, and $p'_{X,k}$ over all $X \in \mathcal{X}$. We store the first
two states $\mathbf{t}_0$ and $\mathbf{t}_1$ losslessly to seed the loop. For each step
$k \geq 2$ the model $\mathcal{F}$ predicts from the two preceding \emph{corrected} states:
\begin{equation}
\mathbf{p}_k = \mathcal{F}(\mathbf{p}'_{k-2},\, \mathbf{p}'_{k-1}).
\end{equation}
For each variable $X$, the residual $t_{X,k} - p_{X,k}$ passes through the same error-bounding  back-end
(Section~\ref{ssec:method_cra5}) to generate the stored residual $q_{X,k}$ and the corrected
slice $p'_{X,k}$. We assemble the corrected slices into $\mathbf{p}'_k$
and feed it forward. Figure~\ref{fig:gc_pred} shows the loop and
Algorithm~\ref{alg:meta-timestamps-selection} lists the steps. The compressed output consists of
the two seed states and the residual stream $\{q_{X,k}\}$. We use {\GraphCast} in
inference mode with the released weights and no fine-tuning.

\begin{algorithm}[tp]
\DontPrintSemicolon
\SetAlgoLined

\KwIn{Full states $\mathbf{t}_0,\dots,\mathbf{t}_{N+1}$ of dataset $\mathcal{X}$;
\newline relative error bound $\epsilon$;
\newline number of prediction steps $N$;
\newline pretrained forecasting model $\mathcal{F}$ (GraphCast or Aurora)}
\KwOut{Compressed bitstream $Z$}

\Comment{Store the two seed states losslessly}
$Z \gets \text{store}(\mathbf{t}_0, \mathbf{t}_1)$\;
$\mathbf{p}'_0 \gets \mathbf{t}_0$;\quad $\mathbf{p}'_1 \gets \mathbf{t}_1$\;

\For{$k \gets 2$ \KwTo $N+1$}{
    \Comment{Predict all variables jointly from the two corrected states}
    $\mathbf{p}_k \gets \mathcal{F}(\mathbf{p}'_{k-2},\, \mathbf{p}'_{k-1})$\;
    \ForEach{variable $X \in \mathcal{X}$}{
        $d_{X,k} \gets t_{X,k} - p_{X,k}$\;
        $\delta_X \gets \epsilon \cdot r_X$\;
        $q_{X,k} \gets \big\lfloor d_{X,k} / \delta_X + 1/2 \big\rfloor$\;
        $\Xi \gets \text{encode}(q_{X,k})$\;
        \Comment{Append the compressed residual to the stream}
        $Z. \text{append}(\Xi)$\;
        $p'_{X,k} \gets p_{X,k} + q_{X,k} \cdot \delta_X$\;
    }
    \Comment{Assemble the corrected state and slide the window}
    $\mathbf{p}'_k \gets \{\, p'_{X,k} : X \in \mathcal{X} \,\}$\;
}
\Return $Z$\;
\caption{Autoregressive error-bounded compression (GraphCast, Aurora).}
\label{alg:meta-timestamps-selection}
\end{algorithm}

\paragraph{Decompression.}
Decompression runs the same pipeline backwards. The decoder loads the seeds $\mathbf{t}_0$ and $\mathbf{t}_1$, then at each
step runs $\mathbf{p}_k = \mathcal{F}(\mathbf{p}'_{k-2}, \mathbf{p}'_{k-1})$, decodes
the residual $q_{X,k}$, and applies the same error correction as in compression
(Section~\ref{sec:background}) to recover $p'_{X,k}$. The decoder re-runs $\mathcal{F}$ to reproduce
each prediction, so the encoder and decoder must compute the same values; any difference
would change the corrected state and violate the bound. This holds because $\mathcal{F}$
runs deterministically and both sides use the same corrected states as input.
{\GraphCast}'s default GPU implementation is not bit-reproducible, so we run it in a deterministic
configuration on CPUs.

\paragraph{Overhead and the compression ratio.}
The compressed size $|Z|$ of an auto-regressive method consists of the two seed states plus the
residual stream. To compute compression ratio $\rho$ we also add the overhead of the model weights. However, in many practical cases the model weights are considered a fixed asset shared across many compression tasks. In such a case we exclude that overhead (see Section~\ref{sec:background}) and calculate compression ratio $\overline{\rho}$. For completeness, in our experimental evaluation we report both quantities.

\paragraph{Stability by construction.}
Long auto-regressive sequences may drift because small errors at each fed-back state can accumulate.
Our approach prevents this. The bound of
Section~\ref{sec:background} keeps every corrected state within $\epsilon\,r_X/2$ of the
truth at each step, so the state fed to the model is always close to the true value. The
correction is applied at every step before the state moves forward, which re-anchors
the sequence to the data, avoiding error accumulation.
Section~\ref{sec:stability} confirms this over the full 1{,}997-step rollout.

\subsection{\Aurora: Auto-regressive Transformer}
\label{ssec:method_aurora}

{\Aurora}~\cite{bodnar2024aurora} is also a temporal forecaster that we integrate into the
same auto-regressive loop as {\GraphCast} (Section~\ref{ssec:method_gc}), with the same
seed states, back-end, and feedback of the corrected state
$\mathbf{p}'_k$. Consequently, the compression pipeline is the one of Figure~\ref{fig:gc_pred}.

\begin{figure}[tp]
    \centering
    \resizebox{\columnwidth}{!}{\begin{tikzpicture}[
  font=\small,
  >=Latex,
  proc/.style ={draw=black!45, line width=0.5pt, rounded corners=3pt, fill=black!4,
                align=center, font=\scriptsize, minimum width=16mm, minimum height=9mm,
                inner sep=2.5pt},
  core/.style ={draw=violet!55!black, line width=0.5pt, rounded corners=3pt,
                fill=violet!12, align=center, font=\scriptsize, minimum width=16mm,
                minimum height=9mm, inner sep=2.5pt},
  flow/.style ={->, line width=0.7pt, black!75},
  tlab/.style ={font=\scriptsize, align=center},
]

\draw[draw=black!30, line width=0.5pt, fill=white] (0.0,-0.15) rectangle (0.6,0.45);
\draw[black!18, line width=0.4pt] (0.0,0.05) -- (0.6,0.05);
\draw[black!18, line width=0.4pt] (0.0,0.25) -- (0.6,0.25);
\draw[black!18, line width=0.4pt] (0.2,-0.15) -- (0.2,0.45);
\draw[black!18, line width=0.4pt] (0.4,-0.15) -- (0.4,0.45);
\draw[draw=black!45, line width=0.5pt, fill=white] (0.15,-0.3) rectangle (0.75,0.3);
\draw[black!22, line width=0.4pt] (0.15,-0.1) -- (0.75,-0.1);
\draw[black!22, line width=0.4pt] (0.15,0.1) -- (0.75,0.1);
\draw[black!22, line width=0.4pt] (0.35,-0.3) -- (0.35,0.3);
\draw[black!22, line width=0.4pt] (0.55,-0.3) -- (0.55,0.3);
\node[tlab] at (0.45,-0.62) {$\mathbf{p}'_{k-2},\,\mathbf{p}'_{k-1}$};

\node[proc] (enc)  at (2.3,0) {3D Perceiver\\encoder};
\node[core] (core) at (4.4,0) {3D Swin\\Transformer\\U-Net};
\node[proc] (dec)  at (6.5,0) {3D Perceiver\\decoder};

\draw[draw=black!45, line width=0.5pt, fill=white] (7.7,-0.3) rectangle (8.3,0.3);
\draw[black!22, line width=0.4pt] (7.7,-0.1) -- (8.3,-0.1);
\draw[black!22, line width=0.4pt] (7.7,0.1) -- (8.3,0.1);
\draw[black!22, line width=0.4pt] (7.9,-0.3) -- (7.9,0.3);
\draw[black!22, line width=0.4pt] (8.1,-0.3) -- (8.1,0.3);
\node[tlab] at (8.0,-0.62) {$\mathbf{p}_k$};

\draw[flow] (0.75,0)    -- (enc.west);
\draw[flow] (enc.east)  -- (core.west);
\draw[flow] (core.east) -- (dec.west);
\draw[flow] (dec.east)  -- (7.7,0);

\end{tikzpicture}}
    \caption{\Aurora{} forecast path. A 3D Perceiver encoder maps the two
    preceding states $\mathbf{p}'_{k-2}, \mathbf{p}'_{k-1}$ to a fixed latent, a 3D
    Swin Transformer U-Net advances it, and a 3D Perceiver decoder produces the
    next state $\mathbf{p}_k$.}
    \label{fig:aurora}
\end{figure}
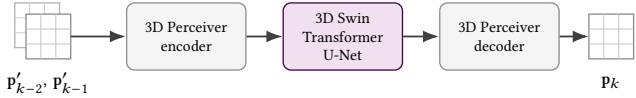

It differs from {\GraphCast} in scale, architecture, and training.
{\Aurora} has 1.3\,B parameters, far more than {\GraphCast}'s 36.7\,M. Its backbone is
a 3D Swin Transformer U-Net with a perceiver-based encoder and decoder that map a
variable set of input fields to a fixed latent representation (see Figure~\ref{fig:aurora}). 
It splits the
global state into patches, takes the two preceding states as input, and processes them
with an attention mechanism. Unlike {\GraphCast}, which is
trained on ERA5 alone, {\Aurora} is pretrained on more data
from six sources (including ERA5), and then fine-tuned. It predicts 4 surface and 5
atmospheric variables on 13 pressure levels (50 to 1000\,hPa), a different native level
set from {\GraphCast}'s 37. These differences change prediction quality, and therefore
compressibility.

\section{Experimental Evaluation}
\label{sec:experiments}

This section evaluates experimentally whether using an elaborate ML model as predictor can improve error-bounded compression. 
We report the primary metric, dataset-level compression ratio (Section~\ref{sec:cr}), the compression and decompression throughput (Section~\ref{sec:throughput}), as well as the reconstruction fidelity of each method (Section~\ref{sec:quality}). We then perform a detailed ablation study that includes per-variable compression, initialization overhead and auto-regression stability (Section~\ref{sec:temporal}). Finally, we study the apparent paradox that more accurate predictions need not compress better, through an analysis of
residual structure (Section~\ref{ssec:residual_structure}).

\subsection{Experimental Setup}
\label{sec:setup}

\subsubsection{Dataset}

Our evaluation uses the ERA5 reanalysis dataset~\cite{hersbach2020era5}, which provides hourly global weather and climate data on a $0.25^{\circ}$ grid of the globe. The grid corresponds to a 2D latitude-longitude matrix with $M = 721$ rows and $N = 1440$ columns. 
As summarized in Table~\ref{table:era5-data}, ERA5 contains variables that fall into two categories: \myNum{i} single-level, such as surface pressure (\texttt{sp}), are 3-dimensional matrices in $\mathbb{R}^{T \times M \times N}$, where $T$ is the number of timesteps; and \myNum{ii} multiple-level variables such as specific humidity (\texttt{Q}), that add an extra dimension for the atmospheric 37 pressure levels (i.e., altitude). In this work, the slice of a multi-level variable at each pressure level is considered a distinct 3D variable and denoted together with its pressure level (e.g., \texttt{Q@500hPa}).    

Each variable $X \in \mathbb{R}^{T \times M \times N}$ is compressed independently, with no cross-variable or cross-level information shared during quantization or entropy coding.

We extract $T=1,997$ consecutive timesteps from the ERA5 dataset at 6-hour resolution, as \GraphCast{} and \Aurora{} operate at this resolution. covering approximately 500 days. Each timestep contains around 4.15\,MB of single-precision floating-point values; the total size of the data is around 1.7\,TB. Not every variable is predicted by every model; Table~\ref{table:era5-data} reports the per-model coverage.  We choose 9 variables common to all methods: \myNum{i} single-level 2m temperature (\texttt{T2m}), 10m $u$- and $v$-wind (\texttt{U10, V10}), and mean sea level pressure (\texttt{MSL}); and \myNum{ii} multi-level temperature (\texttt{T}), geopotential (\texttt{Z}), $u$- and $v$-wind (\texttt{U, V}), and specific humidity (\texttt{Q}).
Following standard weather forecasting practice, we include representative results at the 500 and 850\,hPa pressure levels.

\subsubsection{Hardware} 
\label{sec:hardware}
{\CRA5} and \Aurora{} were run on an NVidia A100-SXM4-80GB GPU with 80GB HBM, installed on a machine with an AMD EPYC 7713P 64-Core CPU @ 2.0GHz and 512GB RAM with Rocky Linux release 9.4.
\SZ{} and \GraphCast{} where run on an Intel Xeon Gold 6246 12-Core CPU @ 3.30GHz, 3TB RAM machine with Rocky Linux release 9.4.
\GraphCast{} was executed on CPU by necessity: its JAX implementation is non-deterministic on GPUs, which is incompatible with the requirements of compression; forcing deterministic GPU execution is no faster than
CPU. 

\begin{table}[t]
\centering

\setlength{\tabcolsep}{4pt}
\caption{ERA5 variables used for evaluation, with short names and which models predict each (\cmark): \GraphCast{} (GC), \Aurora{} (AR), \CRA5{} (CR); MLevel denotes multiple levels.}
\label{table:era5-data}
\begin{tabular}{lccccc}
\toprule
Variable & SName & MLevel & GC & AR & CR \\
\midrule
Geopotential               & Z    & \cmark & \cmark & \cmark & \cmark \\
Specific humidity          & Q    & \cmark & \cmark & \cmark & \cmark \\
Relative humidity          & R    & \cmark &  &  & \cmark \\
Temperature                & T    & \cmark & \cmark & \cmark & \cmark \\
U component of wind         & U    & \cmark & \cmark & \cmark & \cmark \\
V component of wind         & V    & \cmark & \cmark & \cmark & \cmark \\
Vertical velocity          & W    & \cmark & \cmark &  & \cmark \\
2m temperature        & T2m  &  & \cmark & \cmark & \cmark \\
10m u wind component  & U10  &  & \cmark & \cmark & \cmark \\
10m v wind component  & V10  &  & \cmark & \cmark & \cmark \\
100m u wind component & U100 &  &  &  & \cmark \\
100m v wind component & V100 &  &  &  & \cmark \\
Mean sea level pressure    & MSL  &  & \cmark & \cmark & \cmark \\
Total precipitation        & TP   &  & \cmark &  & \cmark \\
Total cloud cover          & TCC  &  &  &  & \cmark \\
Surface pressure           & SP   &  &  &  & \cmark \\
\bottomrule
\end{tabular}
\end{table}

\subsection{Compression Ratio of the Entire Dataset}
\label{sec:cr}

We use three relative error bounds $\epsilon \in \{10^{-2}, 10^{-3}, 10^{-4}\}$ to compress the entire dataset. In Table~\ref{tab:dataset_cr} we show the compression ratio $\rho$. For example, for $\epsilon = 10^{-2}$ \SZ{} achieves $\rho \approx 199$ meaning that it compresses the 1.7~TB dataset down to 8.6~GB.  \SZ{} achieves consistently higher $\rho$ than all ML-based predictors for all error bounds, whereas \CRA5{} is the best among the ML-based methods. As the error bound tightens, all methods converge since the tolerance leaves little prediction-dependent room.

Recall from Section~\ref{sec:background} that the formula for $\rho$ considers the compressed data as well as all overhead; for ML-based methods the model weights are included in the overhead. We claim that in many  cases this is unfair because the ML model is constant; consequently, its cost is amortized among numerous compression tasks. For example, we use only a small extract of the ERA5 dataset, whose total size in ARCO Zarr format is around 5~PB and continues to grow. We can use the same ML models to compress all these data and all future updates. For these reasons, we define compression ratio $\overline{\rho}$ that excludes the overhead of the ML model weights.

Table~\ref{tab:dataset_cr} also shows $\overline{\rho}$, where it is clear that the compression ratio of all ML-based methods is improved. This is particularly evident for \Aurora{}; for instance, for $\epsilon = 10^{-2}$, $\rho \approx 56$ whereas $\overline{\rho} \approx 110$. This happens because its residuals compress so well that the size of the model weights rivals the entire compressed dataset, roughly halving its compression ratio. \CRA5{} and \GraphCast{} are less affected, and the effect fades at
tighter bounds as the growing residuals dominate the fixed cost. Obviously, \SZ{} is not affected because it does not store model weights. For the rest of the paper we will report compression ratio using $\overline{\rho}$.

In summary, none of the ML-based predictors is better than \SZ{}. However this statement does not account for the fact that individual variables behave differently, and does not consider the fidelity of the compressed data. We will investigate these issues below.

\begin{table}[tp]
\centering
\caption{Compression ratio for the entire dataset for relative error bound $\epsilon \in \{10^{-2}, 10^{-3}, 10^{-4}\}$. We report compression ratio $\rho$ that includes all overhead, as well as $\overline{\rho}$ that excludes the size of the ML-model weights.}
\label{tab:dataset_cr}
\begin{tabular}{l rr rr rr}
\toprule
Method & \multicolumn{2}{c}{$\epsilon=10^{-2}$} & \multicolumn{2}{c}{$\epsilon=10^{-3}$} & \multicolumn{2}{c}{$\epsilon=10^{-4}$} \\
\cmidrule(lr){2-3} \cmidrule(lr){4-5} \cmidrule(lr){6-7}
 & $\rho$ & $\overline{\rho}$  & $\rho$ & $\overline{\rho}$  & $\rho$ & $\overline{\rho}$  \\
\midrule
\SZ{}     & \textbf{198.80} & -- & \textbf{36.07} & -- & \textbf{11.72} & --  \\
\CRA5{}      & 130.54 & 136.68 & 19.83 & 19.96 & 7.05 & 7.07 \\
\GraphCast{} & 59.87  & 60.15  & 13.77 & 13.78 & 5.35 & 5.35 \\
\Aurora{}    & 55.87 & 109.72 & 14.35 & 16.41 & 5.20 & 5.45 \\
\bottomrule
\end{tabular}
\end{table}

\subsection{Compression and Decompression Speed}
\label{sec:throughput}

Although this is not the focus of this paper, for completeness we show in Table~\ref{tab:throughput} the end-to-end compression and decompression throughput, measured as MB/s of uncompressed data. 
As expected, \SZ{} is the fastest because of the simple polynomial predictor. \CRA5{} achieves roughly half of \SZ{}'s throughput because the calculation of the forward pass of its autoencoder is more complex. Observe that both auto-regressive methods, \GraphCast{} and \Aurora{} are at least one order of magnitude slower. For \Aurora{} this is expected due to the large model size. \GraphCast{} however, is slow because it is executed on a CPU. Recall from Section~\ref{sec:hardware} that we could not employ a GPU due to the non-deterministic implementation. It must be noted that we did not attempt to optimize any method with regard to execution speed. There are many opportunities for memory and I/O optimization as well as parallelization, but they fall outside the scope of this paper.

\subsection{Reconstruction Quality}
\label{sec:quality}

Although all methods guarantee the relative error bound $\epsilon$, they generate different error distributions that affect the quality of the decompressed data, as shown below. 

\begin{table}[t]
\centering
\caption{Throughput (MB/s of uncompressed data) at $\epsilon = 10^{-2}$, averaged over five timesteps. Rates partly reflect the per-method hardware (Section~\ref{sec:hardware}). }
\label{tab:throughput}
\begin{tabular}{l rr}
\toprule
Method & Compress  (MB/s) & Decompress (MB/s) \\
\midrule
\SZ{}     & \textbf{348.9} & \textbf{1384.8} \\
\CRA5{}      & 237.8 & 723.2 \\
\GraphCast{} & 30.7  & 33.3 \\
\Aurora{}    & 36.1  & 46.0 \\
\bottomrule
\end{tabular}
\end{table}

\subsubsection{Pointwise Error Statistics}

We first ask how much the methods differ in \emph{average} error under a common bound.
Table~\ref{tab:error_stats} reports per-variable MAE and RMSE at $\epsilon = 10^{-2}$.
\Aurora{} improves on the smoothest fields (mean sea level pressure and geopotential),
reducing  MAE by $27$--$34\%$ over \SZ{} by concentrating residuals in the zero bin.
However, in terms of the winds and specific humidity it is not better than \SZ{}. \CRA5 goes further, producing the lowest MAE on $8$ out of $9$ variables,
including a $91\%$ reduction on geopotential, with the exception of specific humidity, which is less structured. The gain is bound-dependent, however: at $\epsilon = 10^{-3}$ \SZ{} has lowest MAE for most variables. (refer to technical report\footnotemark[\value{footnote}])

\begin{table*}[tp]
    \centering
    \setlength{\tabcolsep}{14pt}
    \caption{Mean absolute error (MAE) and root mean square error (RMSE) per variable at $\epsilon = 10^{-2}$, averaged over all 1{,}997 timesteps (atmospheric variables averaged across \Aurora{}'s 13~levels). \textbf{Bold} marks the lowest error per row.}
    \label{tab:error_stats}
    \begin{tabular}{lrrrrrrrr}
    \toprule
    Var & \multicolumn{2}{c}{\SZ} & \multicolumn{2}{c}{\CRA5} & \multicolumn{2}{c}{\GraphCast} & \multicolumn{2}{c}{\Aurora} \\
     & MAE & RMSE & MAE & RMSE & MAE & RMSE & MAE & RMSE \\
    \midrule
    T2m & 4.3e-01 & 5.3e-01 & \textbf{3.6e-01} & \textbf{4.8e-01} & 4.2e-01 & 5.3e-01 & 4.6e-01 & 5.6e-01 \\
    U10 & 2.5e-01 & 3.0e-01 & \textbf{2.1e-01} & \textbf{2.8e-01} & 3.1e-01 & 3.7e-01 & 2.9e-01 & 3.5e-01 \\
    V10 & 2.4e-01 & 3.0e-01 & \textbf{2.0e-01} & \textbf{2.6e-01} & 3.0e-01 & 3.6e-01 & 2.9e-01 & 3.5e-01 \\
    MSL & 4.4e+01 & 5.4e+01 & \textbf{7.2e+00} & \textbf{1.2e+01} & 5.9e+01 & 7.0e+01 & 3.2e+01 & 4.1e+01 \\
    T & 2.4e-01 & 3.0e-01 & \textbf{1.4e-01} & \textbf{1.9e-01} & 2.9e-01 & 3.5e-01 & 2.8e-01 & 3.4e-01 \\
    U & 4.5e-01 & 5.5e-01 & \textbf{3.6e-01} & \textbf{4.7e-01} & 6.1e-01 & 7.2e-01 & 5.8e-01 & 7.0e-01 \\
    V & 4.6e-01 & 5.5e-01 & \textbf{3.2e-01} & \textbf{4.3e-01} & 6.1e-01 & 7.3e-01 & 5.9e-01 & 7.1e-01 \\
    Q & \textbf{3.6e-05} & \textbf{4.4e-05} & 4.0e-05 & 4.9e-05 & 4.3e-05 & 5.1e-05 & 3.8e-05 & 4.8e-05 \\
    Z & 5.6e+01 & 6.8e+01 & \textbf{5.2e+00} & \textbf{7.4e+00} & 8.3e+01 & 9.8e+01 & 3.7e+01 & 4.7e+01 \\
    \bottomrule
    \end{tabular}
\end{table*}

\subsubsection{Error Distribution Analysis}

\begin{figure*}[tp]
\centering
\includegraphics[width=\textwidth]{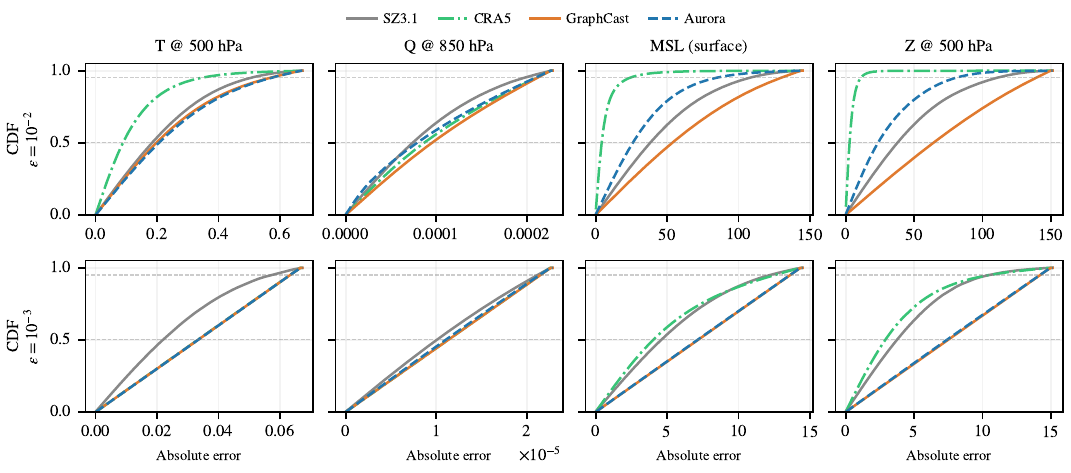}
\caption{
  CDFs of pointwise absolute error at
  $\epsilon = 10^{-2}$ (top) and $\epsilon = 10^{-3}$ (bottom) for four representative
  variables; a steeper rise means error concentrated nearer zero. Dashed lines mark the
  median and 95th percentile.
}
\label{fig:cdf}
\end{figure*}

Figure~\ref{fig:cdf} shows the CDF of pointwise absolute error over all timesteps. At $\epsilon = 10^{-2}$ the ordering is
variable-dependent and tracks the fidelity results: \Aurora{} rises fastest for the smooth
fields (\texttt{MSL}, geopotential) but not for temperature or humidity, while \CRA5 leads on all
but specific humidity. At $\epsilon = 10^{-3}$ the picture reverses: \SZ{} now rises fastest for every variable, with only \CRA5{} still
ahead on the smooth fields. This crossover is a central finding: temporal prediction
sharpens the error distribution at moderate tolerances but not at tight ones, where
narrow quantization bins let \SZ{}'s spatially coherent errors quantize more efficiently
than the scattered errors of the temporal models. The practical consequence is
application-dependent: learned prediction gives better aggregate fidelity for predictable
fields at moderate bounds, whereas at tight bounds \SZ{} wins on both compression ratio
and average error.

\subsubsection{Spatial Error Structure}

\begin{figure*}[tp]
\centering
\includegraphics[width=\textwidth]{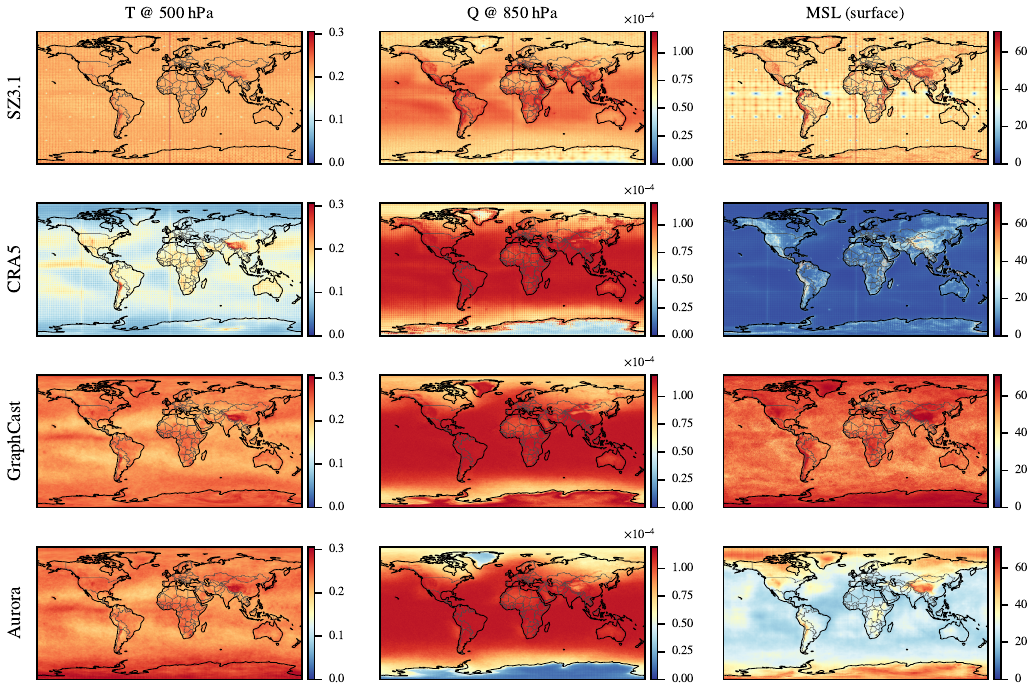}
\caption{%
  Time-averaged per-pixel MAE at $\epsilon = 10^{-2}$.
  Rows: \Aurora{}, \GraphCast{}, \SZ{}, \CRA5{}.
  Columns: temperature (500\,hPa), specific humidity (850\,hPa), mean sea level pressure.
  Color scale shared within each column.%
}
\label{fig:pixel_mae}
 \end{figure*}

Figure~\ref{fig:pixel_mae} maps the time-averaged
per-pixel MAE at $\epsilon = 10^{-2}$. The temporal predictors err in step with local
predictability: for temperature the error concentrates along the mid-latitude storm
tracks ($40$--$60^{\circ}$\,N/S) while the tropics stay near-zero, and \Aurora{}'s \texttt{MSL} error
is low across the tropics but rises at the high latitudes of deep cyclones. \SZ{} is
qualitatively different: its error is spatially uniform and uncorrelated with any weather
feature. \CRA5 differs from both: its error
follows sharp geographic gradients (coastlines, land-sea contrasts) rather than weather. 
\SZ{} gives better pointwise fidelity but introduces local artifacts, especially in \texttt{T} and \texttt{MSL}, so it is not always spatially smooth. While forecast-model reconstructions give worse pointwise accuracy but smoother errors and more meteorologically coherent.

\subsection{In-Depth Analysis: Temporal Predictors}
\label{sec:temporal}

Temporal prediction does not, on its own, outperform \emph{spatial} polynomial prediction for overall compression. Below we analyze the per-variable and per-level compression ratio, initialization overhead, and stability with \SZ{} as the baseline. \CRA5 is excluded: reconstructing each timestep independently, it has no overhead, no feedback, and no per-variable or per-level residual breakdown.

\subsubsection{Per-Variable Compression Ratio}

\begin{figure}[tp]
\centering
\includegraphics[width=.78\columnwidth]{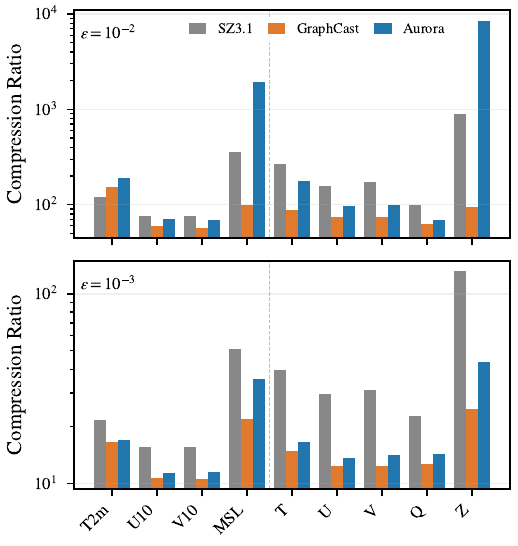}
\caption{
  Per-variable compression ratio at $\epsilon = 10^{-2}$ (top) and $\epsilon = 10^{-3}$ (bottom), excluding initialization overhead; log-scale $y$-axis. \GraphCast{} and \SZ{} use \Aurora{}'s 13~levels.
  }
\label{fig:pervariable_cr}
\end{figure}

In Figure~\ref{fig:pervariable_cr} we present the per-variable compression ratio, excluding the initialization overhead. At $\epsilon = 10^{-2}$ \Aurora{} wins on the smoothest fields: geopotential compresses about $8{,}540\times$ against \SZ{}'s $888\times$, with comparable gains for mean sea level pressure and 2-meter temperature, because near-zero residuals leave quantized streams dominated by zeros. On the more turbulent fields, atmospheric temperature, the winds, and specific humidity, \SZ{} keeps the advantage, as their fine-scale variability resists grid-scale prediction. \GraphCast{} trails both on every field except 2-meter temperature, likely because its mesh-to-grid interpolation injects high-frequency artifacts.
At $\epsilon = 10^{-3}$ the advantage disappears and \SZ{} leads every variable.
As established in Section~\ref{sec:quality}, the benefit of small residuals is confined to moderate tolerances, since tight bins span a wide integer range regardless of residual magnitude.

\subsubsection{Per-Level Compression Ratio}

\begin{figure}[tp]
\centering
\includegraphics[width=\columnwidth]{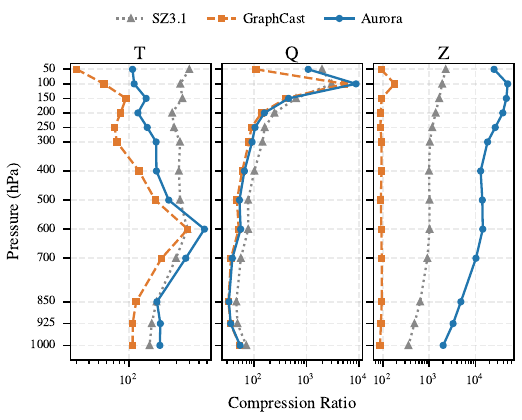}
\caption{
  Per-level compression ratio for temperature, specific humidity, and geopotential at $\epsilon = 10^{-2}$ (\Aurora{}'s 13~levels), excluding initialization overhead.}
\label{fig:perlevel_cr}
\end{figure}

We  ask how this varies with altitude (Figure~\ref{fig:perlevel_cr}). For temperature, \Aurora{} is best in the smooth mid-troposphere, peaking near $580\times$ at 600\,hPa, but is overtaken by \SZ{} above 300\,hPa and matched by it below 850\,hPa. Humidity compresses well for every method at the near-dry upper levels (50--100\,hPa) and drops sharply below 300\,hPa, where \SZ{} leads.

Geopotential is the most extreme: \Aurora{} climbs from about $670\times$ near the surface to over $45{,}000\times$ at the top of the atmosphere, where its predictions are nearly perfect. These per-level peaks do not survive aggregation, however: the variable total is dominated by the lower-tropospheric levels, which carry the largest compressed sizes and thus most of the bytes (a harmonic-mean effect).

\subsubsection{Initialization Overhead and Amortization}
\label{sec:amortization}
\begin{figure*}[tp]
\centering
\includegraphics[width=\textwidth]{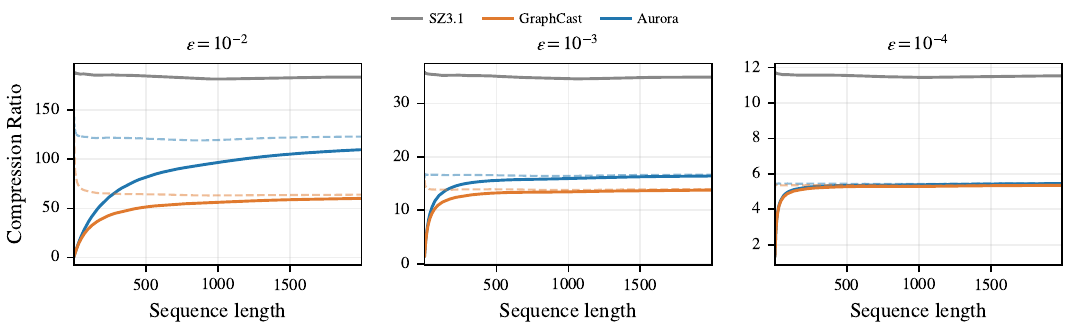}
\caption{ 
  Compression ratio versus sequence length for all three methods
and error bounds. Solid lines include the two-timestep initialization overhead in the
denominator; dashed lines give the asymptotic CR without it (\SZ{} has none).}
\label{fig:cr_amortization}
\end{figure*}

The temporal methods store two uncompressed seed timesteps, so we check whether this fixed overhead materially lowers their ratios (Figure~\ref{fig:cr_amortization}). Convergence to the asymptotic ratio depends sharply on the bound. When residuals are large ($\epsilon = 10^{-4}$), the overhead becomes negligible within roughly 80 timesteps. At $\epsilon = 10^{-2}$, by contrast, the residuals compress so well that the two seed frames still dominate the stored size after the full 1{,}997-step sequence.

Over the multi-year archives these methods target, the overhead amortizes at every bound; within our window it depresses the dataset-level ratio only at $\epsilon = 10^{-2}$. We therefore report dataset-level ratios (Table~\ref{tab:dataset_cr}) with the overhead included and the per-variable and per-level figures (Figures~\ref{fig:pervariable_cr},~\ref{fig:perlevel_cr}) without it, isolating intrinsic residual compressibility.

\subsubsection{Autoregressive Stability}
\label{sec:stability}

\begin{figure}[tp]
\centering
\includegraphics[width=\columnwidth]{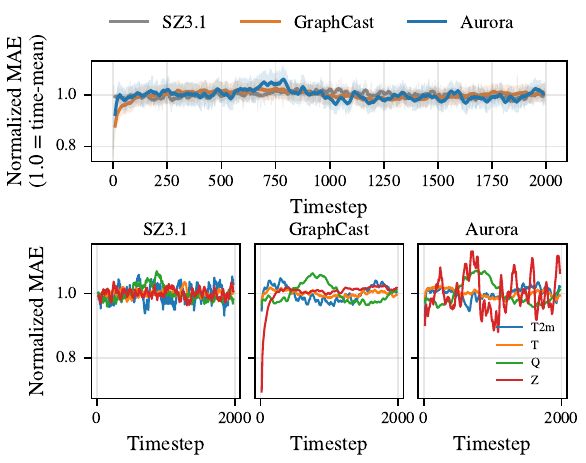}
\caption{Normalized MAE over the 1{,}997-step autoregressive rollout at $\epsilon = 10^{-2}$; each variable's MAE is divided by its time-mean. \textbf{Top}: per-method average over all variables (bold: 20-step rolling average; shaded: raw values). \textbf{Bottom}: per-variable curves for each method (representative variables).}
\label{fig:error_accum}
\end{figure}

A natural concern is whether quantization artifacts compound through the feedback loop: in free-running forecasting, these models diverge after roughly 10~days ($\sim$40 steps).
Figure~\ref{fig:error_accum} tracks the normalized MAE over the full 1{,}997-step sequence ($\sim$500 days) at $\epsilon = 10^{-2}$: all three methods stay stationary around 1.0 with no upward trend. The closed loop guarantees this, feeding each prediction from \emph{decompressed} states within $\delta_v$ of the truth so the input cannot drift far from reality. The $\pm 5\%$ oscillations reflect seasonal variation in predictability, not drift; \SZ{}'s are marginally larger because its error tracks each timestep's spatial complexity, whereas the temporal predictors' curves are smoother. The rollout is therefore stable over the entire sequence, so the error-bounded feedback removes the drift that limits free-running forecasts.

\subsection{Residual-Structure Analysis}
\label{ssec:residual_structure}

The learned predictors often reconstruct  better than \SZ{}, yet do not yield a higher
compression ratio. To explain this, note that with the shared back-end the ratio is fixed by the
quantized-residual stream, not by reconstruction accuracy: at a fixed error bound,
$\mathrm{CR}=\frac{B_0}{R}$ and $R \approx H(Q)$, 
where $B_0$ is the raw bit-width per value, $R$ the achieved bits per value, and $H(Q)$ the symbol
entropy of the quantized residual, the floor the Huffman stage works against. Compressibility is
therefore governed by residual \emph{entropy}, and a more accurate prediction need not yield a
lower-entropy residual.

Table~\ref{tab:residual_structure} separates two error measures that are usually conflated, for two
representative fields (a smooth surface field and a turbulent mid-tropospheric one).
$\mathrm{MAE_{rec}}$ is the \emph{reconstruction} error $|\mathrm{truth}-\mathrm{recon}|$, the
fidelity the user observes, and averages over \emph{all} points, so it is dominated by the
well-predicted majority. $\mathrm{MAE_{res}}$ is the \emph{residual} magnitude $|2q\epsilon|$, the
mean size of the corrections actually stored; it depends on how \emph{many} points exceed the bound
(\emph{zero \%} counts those within it) and how large their corrections are. $\mathrm{MAE_{res}}$,
not fidelity, tracks $H(Q)$ and hence the rate. The two measures can move in \emph{opposite}
directions: a predictor can be excellent on the bulk (low $\mathrm{MAE_{rec}}$) yet correct enough
outliers to inflate the stored residual, which is precisely what costs bits.

\begin{table}[tp]
\centering
\setlength{\tabcolsep}{4pt}
\caption{Reconstruction error, residual magnitude, zero-bin fraction, entropy $H(Q)$, and
compression ratio for two representative fields ($\epsilon=10^{-2}$, $200$ timesteps). \textbf{Bold} marks cases discussed in the text.}
\label{tab:residual_structure}
\begin{tabular}{@{}l l r r r r r@{}}
\toprule
Var & Method & $\mathrm{MAE_{rec}}$ & $\mathrm{MAE_{res}}$ & zero \% & $H(Q)$ & CR \\
\midrule
\multirow{4}{*}{T2m}
 & \SZ{}      & 4.3e-01 & 1.2e-01 & $96.2$ & $0.3$ & $117.4$ \\
 & \CRA5      & \textbf{3.6e-01} & \textbf{2.1e-01} & $92.4$ & $\mathbf{0.5}$ & $\mathbf{103.5}$ \\
 & \GraphCast & 4.3e-01 & 1.4e-01 & $94.6$ & $0.4$ & $145.9$ \\
 & \Aurora    & 4.6e-01 & 1.2e-01 & $95.4$ & $0.3$ & $177.8$ \\
\midrule
\multirow{4}{*}{Q@850}
 & \SZ{}      & \textbf{8.3e-05} & \textbf{7.9e-05} & $88.6$ & $\mathbf{0.8}$ & $\mathbf{47.9}$ \\
 & \CRA5      & 9.4e-05 & 1.4e-04 & $74.2$ & $1.3$ & $34.0$ \\
 & \GraphCast & 1.0e-04 & 1.7e-04 & $71.2$ & $1.5$ & $33.4$ \\
 & \Aurora    & \textbf{8.9e-05} & \textbf{1.9e-04} & $70.5$ & $\mathbf{1.5}$ & $34.0$ \\
\bottomrule
\end{tabular}
\end{table}

\paragraph{Better fidelity need not lower the entropy.}
On \texttt{T2m}, \CRA5{} has the \emph{best} reconstruction MAE ($0.36$ vs.\ \SZ{}'s $0.43$)
but the \emph{largest} residual MAE ($0.21$ vs.\ $0.12$): it predicts most points accurately yet
leaves a correction at more of them ($7.6\%$ outside the bound vs.\ $3.8\%$), so its residual
entropy is the highest ($0.5$ vs.\ $0.3$ bits) and its CR the lowest ($103.5$ vs.\ $117.4$). On
$850$\,hPa specific humidity, \Aurora{} matches \SZ{}'s fidelity ($8.9$ vs.\ $8.3\times10^{-5}$) yet
stores a residual $2.4\times$ larger ($1.9$ vs.\ $0.8\times10^{-4}$), with entropy nearly double
($1.5$ vs.\ $0.8$ bits), corresponding to corrections at $30\%$ of points vs.\ $11\%$, so its CR is
$34.0$ vs.\ $47.9$. Equal-or-better fidelity yields a larger stored residual and lower compression.

\paragraph{Consequence for the dataset.}
The learned methods' fidelity advantage therefore does not become a dataset-level CR gain. The aggregate ratio is
byte-weighted toward the turbulent, high-entropy fields (humidity, winds), where their residuals carry
up to twice \SZ{}'s entropy, while the smooth fields on which they win (geopotential, \texttt{MSL}) contribute
little. The advantage is also mainly \CRA5{}'s: \Aurora{} and \GraphCast{} do not perform better than \SZ{} on the
mid-tropospheric and turbulent fields.

\begin{figure}[tp]
\centering
\includegraphics[width=\columnwidth]{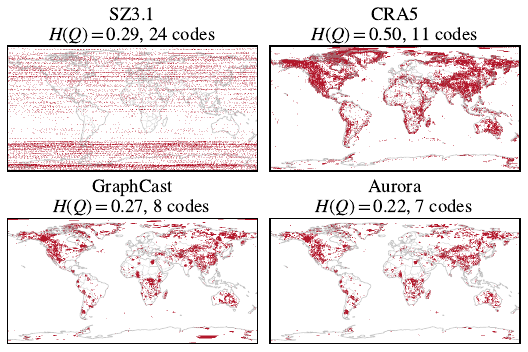}
\caption{Binary quantized-residual support for \texttt{T2m} at one timestep ($\epsilon=10^{-2}$):
red marks where a correction is stored, white where the prediction is within the bound.}
\label{fig:residual_map}
\end{figure}

\paragraph{Where the corrections occur.}
Figure~\ref{fig:residual_map} maps the points at which each method stores a correction. \SZ{}'s form
a spatially \emph{unstructured} scatter of isolated pixels, whereas every learned method concentrates its
corrections into coherent regions following coastlines and weather features. This matters because
the back-end is order-0 (Huffman prices each symbol by its global frequency) followed by a
one-dimensional run/dictionary stage (Zstandard, which exploits only exact repeats along the
row-major scan), so the \emph{two-dimensional} structure of the learned residuals is largely invisible to
it. A predictor whose errors are spatially organized leaves compression unrealized under a
conventional 1-D pipeline, which points to spatially-aware (2-D) residual coding as a route to
convert the learned methods' prediction quality into compression ratio.
\section{Related Work}\label{sec:Related Work}

\subsection{Classical Error-Bounded Lossy Compression}
\label{ssec:rw_classical}
Among error-bounded compressors, ZFP~\cite{lindstrom2014fixed} operates in fixed-rate 
mode by truncating coefficients after an orthogonal block transform, while 
SPERR~\cite{li2023lossy} applies the SPECK wavelet 
algorithm~\cite{pearlman2004efficient,tang2006three} to structured scientific grids. 
The SZ family represents the most widely adopted class of such compressors. SZ2+ 
~\cite{tao2017significantly} introduced multidimensional Lorenzo predictors; 
SZ3~\cite{liang2022sz3} and SZauto~\cite{zhao2020significantly} extended the 
prediction and quantization pipeline; and SZ3.1~\cite{zhao2021optimizing} incorporated 
dynamic multidimensional spline interpolation. Recently, HPEZ~\cite{liu2023high} 
added natural cubic splines, multidimensional interpolation strategies, and 
quality-driven auto-tuning. All of these methods predict from spatially 
neighboring points within a single timestep, leaving temporal redundancy not entirely 
exploited. We evaluate against SZ3.1~\cite{zhao2021optimizing} as the sole classical baseline. 
Although HPEZ~\cite{liu2023high} is more recent and extends SZ3.1 with additional 
interpolation strategies, our preliminary experiments on a small subset of the data 
showed that SZ3.1 matches or closely approaches HPEZ in compression ratio across 
several tested variables and error bounds, with any gains from HPEZ remaining marginal.

\subsection{ML-based Lossy Compression}
\label{ssec:rw_neural}

\paragraph{INR: Implicit Neural Representations-based Compression}
CoordNet~\cite{han2022coordnet} represents time-varying volumetric fields as a 
coordinate-based MLP mapping $(x,y,z,t)$ to scalar values; the network weights 
serve as the compressed representation, enabling continuous queries and high 
compression ratios. KD-INR\cite{han2023kd} addresses the scalability problem of single-model INRs 
over long sequences by training compact per-timestep teacher INRs and distilling 
them into a single student network with temporal embeddings, reducing cross-time 
redundancy without sacrificing per-timestep detail. COIN++~\cite{dupont2022coin} meta-learns a 
shared INR backbone and compresses new instances by optimizing only small FiLM-style 
modulations per patch, drastically reducing encoding cost while maintaining quality 
across diverse modalities including climate fields. Huang et 
al.~\cite{huang2023compressing} take a similar approach specifically for weather and 
climate data, overfitting coordinate networks to ERA5 fields and storing quantized, 
entropy-coded weights as the compressed representation.

\paragraph{Autoencoder-based Compression}
HAE \cite{li2024attention} proposes a hierarchical autoencoder in which a coarse-level AE captures global structure, and successive finer-level AEs encode residuals at increasing spatial 
detail; each level's latents are independently quantized and entropy-coded, allowing 
rate-distortion tradeoffs. 

Hayne et 
al.~\cite{hayne2021using} applied a variational autoencoder~\cite{balle2018variational} 
to 2D floating-point scientific fields, achieving high compression ratios without 
error guarantees. 

\paragraph{Hybrid methods with error bounds}
AE-SZ~\cite{liu2021exploring} integrates a learned convolutional AE predictor with 
the classical Lorenzo predictor inside the SZ framework. 
SRN-SZ~\cite{liu2023srn} reframes compression as a downsample-then-super-resolve 
pipeline: the encoder stores a coarsened grid, and a high-accuracy super-resolution 
network (HAT) reconstructs fine spatial detail at decode time, falling back to 
adaptive interpolation when the resolution gap is too large. GraphComp~\cite{li2026extreme} maps each timestep to a graph of segmented regions and trains a temporal graph autoencoder whose latent compresses the sequence. 

LLMComp~\cite{li2025llmcomp} casts
compression as auto-regressive token modeling, training a decoder-only transformer that
stores a top-$k$ rank index when it predicts the next token and a fallback correction
otherwise. These hybrid approaches 
demonstrate that learned predictors can improve compression ratio at fixed error 
bounds, but all remain domain-agnostic and some do not explicitly exploit the specific spatiotemporal structure of atmospheric data. 

We do not evaluate any of the mentioned autoencoder-based or hybrid methods because: 
\myNum{i} they require training from scratch on the ERA5 data, which is a substantial undertaking that defies the purpose of this work; 
\myNum{ii} for AE-SZ and SRN-SZ the code is not publicly available. We also note that HPEZ
and SZ3.1 represent the state-of-the-art amongst all SZ-based works. Therefore, evaluating against SZ3.1  provides a strong baseline for comparison.

\subsection{Neural Compression for ERA5}
\label{ssec:rw_era5}

We have already discussed CRA5~\cite{han2023cra5} in the previous sections. 

Another neural compressor for ERA5 proposed by Huang et al.~\cite{huang2023compressing} takes a hierarchical neural approach targeting 
the full ERA5 variable set with similar goals. Both are purpose-built ERA5 codecs with strong results, but their standard
evaluation treats the full codec as a monolithic system operating outside an
error-bounded framework.
Huang et al.~\cite{huang2023compressing} is not evaluated here because there is no publicly available pre-trained model and training from scratch is required. Therefore, its integration is left for future work.

\subsection{Weather Forecasting ML Models}
\label{ssec:rw_selection}

Beyond GraphCast~\cite{lam2023learning} and Aurora~\cite{bodnar2024aurora} that have been discussed before, a broader landscape of data-driven 
forecasters has been proposed. Pangu-Weather~\cite{bi2023pangu} 
uses a 3D Earth-Specific Transformer with latitude-dependent positional biases and 
separate model instances for different lead times to limit auto-regressive error 
accumulation. FourCastNet~\cite{pathak2022fourcastnet} applies Adaptive Fourier 
Neural Operators that learn spectral filters in the frequency domain, enabling global 
receptive fields with favorable computational scaling. FengWu\cite{chen2023fengwu} extends this family 
with multi-lead supervision and uncertainty-aware losses targeting skill horizons 
beyond 10 days. GenCast~\cite{price2023gencast} introduces a diffusion-based probabilistic forecaster 
on a spherical mesh, generating calibrated forecast ensembles that improve 
representation of extreme events. Aardvark~\cite{vaughan2024aardvark} replaces the full NWP inference 
stack with a trainable encoder that ingests raw observations, enabling end-to-end 
optimization without NWP inputs at deployment time.

These models are excluded from our evaluation due to limited 
variable coverage, less accessible inference pipelines, or 
architectural choices that complicate integration into a deterministic 
prediction-residual framework (e.g., GenCast's probabilistic diffusion outputs). We leave their integration for future work.

\section{Limitations and Future Directions}
\label{sec:limitations}

Our study identifies three main limitations of using ML-based predictors for bounded lossy compression: 
\myNum{i} \emph{Computational and storage cost.}
Running a deep neural network in place of a polynomial predictor is computationally costly; the
pretrained weights are also a large fixed side cost stored wherever data is decoded. 
\myNum{ii} \emph{Sequential decoding.}
For auto-regressive methods, reconstructing timestep $k$ requires
rolling the model forward from the seed states through every step in between.
Note that {\CRA5} does not suffer from this problem.
\myNum{iii} \emph{Compression ratio.}
In general, classical compressors achieve higher compression ratio, although ML-based methods can be better for specific variables and may achieve higher fidelity.

There are several opportunities for extending this work: 
\myNum{i} \emph{Additional ML-models.}
Extend the evaluation to include more forecasting models, such as Pangu-Weather, FourCastNet and FengWu.
\myNum{ii} \emph{Structure-aware residual coding.}
Encode the residual to exploit its spatial organization through spatially adaptive quantization.

\myNum{iii}
\emph{Compression-aware predictors.}
Train predictors with a  residual-entropy objective, instead of mean squared error, to match the requirements of the Huffman encoder.

\section{Conclusion}
\label{sec:conclusion}

In this Experimental Analysis and Benchmark (EA\&B) paper, we present the first systematic evaluation of deep learning weather models as predictors
in an error-bounded compression pipeline for ERA5: a learned spatial codec
({\CRA5}) and two temporal forecasters {\Aurora}, {\GraphCast} against the SZ3.1 baseline.
We propose an auto-regressive framework that bounds error at every step and avoids error accumulation.
The headline result is negative: no learned predictor improves dataset-level compression, and SZ3.1 leads at every bound. Yet the ML-based predictors may offer higher fidelity and higher compression ratio for smooth variables. We analyze why higher fidelity does not translate to better compression: the ratio is determined by residual entropy, not accuracy, and the learned residuals are spatially structured in a way the one-dimensional coder cannot exploit. 
We argue that our study is a sound proxy that exposes the bottlenecks of using learned predictors for the general case of scientific data compression.

\begin{acks}
For computer time, we used IBEX and Shaheen III, managed by
the Supercomputing Core Laboratory at KAUST, Saudi Arabia.
\end{acks}


\clearpage
\bibliographystyle{ACM-Reference-Format}
\bibliography{references}

\appendix
\begin{table*}[tp]
\centering
\caption{DL prediction and compression models considered in this survey, organized by 
scope. Selection criteria for our evaluation are recency, code availability, and 
native applicability to ERA5 atmospheric data.}
\label{table:cmp_methods}
\begin{tabular}{llccc}
\hline
\bf Model & \bf Scope & \bf Code Available & \bf ERA5 Native \\
\hline
CoordNet      & Time-Varying Volumes  & \cmark & \xmark \\
KD-INR        & Time-Varying Volumes  & \xmark & \xmark \\
HAE           & General               & \cmark & \xmark \\
ATN-AE        & General               & \xmark & \xmark  \\
GraphComp     & General               & \cmark & \xmark \\
AE-SZ         & General               & \xmark & \xmark \\
SRN-SZ        & General               & \xmark & \xmark \\
COIN++        & General               & \cmark & \xmark \\
CRA5          & Weather (ERA5)        & \cmark & \cmark \\
HH-NN         & Weather (ERA5)        & \cmark & \cmark \\
GraphCast     & Weather-Forecasting   & \cmark & \cmark \\
GenCast       & Weather-Forecasting   & \cmark & \cmark \\
Aurora        & Weather-Forecasting   & \cmark & \cmark \\
Aardvark      & Weather-Forecasting   & \cmark & \cmark \\
Pangu-Weather & Weather-Forecasting   & \cmark & \cmark \\
FengWu        & Weather-Forecasting   & \cmark & \cmark \\
FourCastNet   & Weather-Forecasting   & \cmark & \cmark \\
\hline
\end{tabular}
\end{table*}

\begin{table*}[tp]

    \centering

    \setlength{\tabcolsep}{14pt}

    \caption{Mean absolute error (MAE) and root mean square error (RMSE) per variable at $\epsilon = 10^{-3}$, averaged over all timesteps. \textbf{Bold} marks the lowest error per row.}

    \label{tab:error_stats_eps_1e-3}

    \begin{tabular}{lrrrrrrrr}

    \toprule

    Var & \multicolumn{2}{c}{\SZ} & \multicolumn{2}{c}{\CRA5} & \multicolumn{2}{c}{\GraphCast} & \multicolumn{2}{c}{\Aurora} \\

     & MAE & RMSE & MAE & RMSE & MAE & RMSE & MAE & RMSE \\

    \midrule

    T2m & \textbf{5.7e-02} & \textbf{6.8e-02} & 6.5e-02 & 7.5e-02 & 6.5e-02 & 7.5e-02 & 6.5e-02 & 7.5e-02 \\

    U10 & \textbf{3.3e-02} & \textbf{3.9e-02} & 3.5e-02 & 4.1e-02 & 3.5e-02 & 4.1e-02 & 3.5e-02 & 4.1e-02 \\

    V10 & \textbf{3.2e-02} & \textbf{3.7e-02} & 3.4e-02 & 3.9e-02 & 3.4e-02 & 3.9e-02 & 3.4e-02 & 3.9e-02 \\

    MSL & 5.2e+00 & 6.3e+00 & \textbf{4.9e+00} & \textbf{6.2e+00} & 7.2e+00 & 8.3e+00 & 7.1e+00 & 8.3e+00 \\

    T & \textbf{3.0e-02} & \textbf{3.6e-02} & 3.8e-02 & 4.4e-02 & 3.8e-02 & 4.4e-02 & 3.8e-02 & 4.4e-02 \\

    U & \textbf{5.7e-02} & \textbf{6.8e-02} & 7.0e-02 & 8.1e-02 & 7.0e-02 & 8.1e-02 & 7.0e-02 & 8.1e-02 \\

    V & \textbf{5.7e-02} & \textbf{6.9e-02} & 7.0e-02 & 8.1e-02 & 7.1e-02 & 8.2e-02 & 7.1e-02 & 8.2e-02 \\

    Q & \textbf{4.5e-06} & \textbf{5.4e-06} & 5.1e-06 & 5.9e-06 & 5.1e-06 & 5.9e-06 & 5.0e-06 & 5.8e-06 \\

    Z & 5.4e+00 & 6.6e+00 & \textbf{4.8e+00} & \textbf{6.2e+00} & 9.3e+00 & 1.1e+01 & 9.2e+00 & 1.1e+01 \\

    \bottomrule

    \end{tabular}

\end{table*}

\section{Appendix}
\subsection{Summary of Related Work}
Table~\ref{table:cmp_methods} summarizes the landscape of methods surveyed above 
against three selection criteria: code availability, native ERA5 applicability, and 
inclusion in our evaluation. We focus on pretrained DL forecasting models rather 
than purpose-built compression codecs for the reasons established in 
Section~\ref{sec:intro}: training a compression-specific DL model from scratch 
requires substantial resources, and it is important to first establish whether 
DL-based temporal prediction offers a genuine advantage over classical spatial 
prediction before incurring that cost. Pretrained forecasting models serve as a 
principled proxy for this feasibility study.

\subsection{Error Statistics for Tighter Bound}
Table~\ref{tab:error_stats_eps_1e-3} reports per-variable MAE and RMSE at $\epsilon = 10^{-3}$.
At this tighter error bound, the advantage shifts back to \SZ{}: it achieves the lowest MAE and RMSE on $7$ out of $9$ variables, including near-surface temperature, winds, atmospheric temperature, wind fields, and specific humidity.
The learned residual methods provide little benefit for these less structured or higher-frequency fields, where their residuals are not concentrated enough to offset the additional correction error.
\CRA5{} remains beneficial only for the smoothest large-scale fields, reducing MAE by about $5\%$ on mean sea level pressure and $12\%$ on geopotential relative to \SZ{}.
In contrast, \GraphCast{} and \Aurora{} do not improve over \SZ{} at this bound, and are especially worse on mean sea level pressure and geopotential.
This confirms that the gains observed at $\epsilon = 10^{-2}$ are strongly bound-dependent: when the tolerance is tightened, \SZ{} already preserves most variables accurately, leaving less exploitable residual structure for forecast-based correction.


\end{document}